\documentclass[10pt,journal,compsoc]{IEEEtran}

\ifCLASSOPTIONcompsoc
  \usepackage[nocompress]{cite}
\else
  \usepackage{cite}
\fi

\ifCLASSINFOpdf
  \usepackage[pdftex]{graphicx}
\else
\fi

\usepackage{amsmath}

\usepackage{array}

\ifCLASSOPTIONcompsoc
  \usepackage[caption=false,font=footnotesize,labelfont=sf,textfont=sf]{subfig}
\else
  \usepackage[caption=false,font=footnotesize]{subfig}
\fi

\usepackage{stfloats}

\ifCLASSOPTIONcaptionsoff
  \usepackage[nomarkers]{endfloat}
 \let\MYoriglatexcaption\caption
 \renewcommand{\caption}[2][\relax]{\MYoriglatexcaption[#2]{#2}}
\fi

\usepackage{url}

\usepackage{amssymb,xcolor}
\usepackage{multicol}
\usepackage{multirow}
\usepackage{array,microtype}
\usepackage{tabularx}
\usepackage{enumitem}

\usepackage{placeins}
\usepackage{stfloats}

\newcommand{\mname}[0]{\mbox{{\bf SCG-OL}}}
\newcommand{\newmname}[0]{\mbox{\bf D-SCG-OL}}
\newcommand{\newmnamefull}[0]{{\bf \mbox{D-SCG} Object Localiser}}
\newcommand{\oldgraph}[0]{\mbox{\bf SCG}}
\newcommand{\newgraph}[0]{\mbox{\bf D-SCG}}
\newcommand{\newsg}[0]{\mbox{\bf D-SG}}

\newcolumntype{P}[1]{>{\centering\arraybackslash}p{#1}}
\newcolumntype{C}{>{\centering\arraybackslash}X}

\begin{document}

\title{Leveraging commonsense for object localisation in partial scenes}

\author{Francesco~Giuliari, Geri~Skenderi, Marco~Cristani, Alessio~Del~Bue and Yiming~Wang%
\IEEEcompsocitemizethanks{\IEEEcompsocthanksitem F. Giuliari is with Istituto Italiano di Tecnologia (IIT) and University of Genoa (UniGe)%
\IEEEcompsocthanksitem G. Skenderi is with University of Verona (UniVr)%
\IEEEcompsocthanksitem M. Cristani is with University of Verona (UniVr) and Istituto Italiano di Tecnologia (IIT)%
\IEEEcompsocthanksitem A. Del Bue is with Istituto Italiano di Tecnologia (IIT)%
\IEEEcompsocthanksitem Y. Wang is with Fondazione Bruno Kessler (FBK) and Istituto Italiano di Tecnologia (IIT)%
}%
\thanks{This project has received funding from the European Union’s Horizon 2020 research and innovation programme ``MEMEX'' under grant agreement No 870743, and the Italian Ministry of Education, Universities and Research (MIUR) through PRIN 2017 - Project Grant 20172BH297: I-MALL and ``Dipartimenti di Eccellenza 2018-2022".}
}

\IEEEtitleabstractindextext{%

\begin{abstract}
    We propose an end-to-end solution to address the problem of object localisation in partial scenes, where we aim to estimate the position of an object in an unknown area given only a partial 3D scan of the scene. We propose a novel scene representation to facilitate the geometric reasoning, Directed Spatial Commonsense Graph (D-SCG), a spatial scene graph that is enriched with additional concept nodes from a commonsense knowledge base. Specifically, the nodes of D-SCG represent the scene objects and the edges are their relative positions. Each object node is then connected via different commonsense relationships to a set of concept nodes. With the proposed graph-based scene representation, we estimate the unknown position of the target object using a Graph Neural Network that implements a novel attentional message passing mechanism. The network first predicts the relative positions between the target object and each visible object by learning a rich representation of the objects via aggregating both the object nodes and the concept nodes in D-SCG. These relative positions then are merged to obtain the final position. We evaluate our method using Partial ScanNet, improving the state-of-the-art by 5.9\% in terms of the localisation accuracy at a 8x faster training speed.
\end{abstract}

\begin{IEEEkeywords}
Vision and Scene Understanding, Scene Analysis, Computer vision, Machine learning
\end{IEEEkeywords}}

\maketitle

\IEEEdisplaynontitleabstractindextext

\IEEEpeerreviewmaketitle

\begin{figure}[t!]
	\centering
	\includegraphics[width=0.9\linewidth]{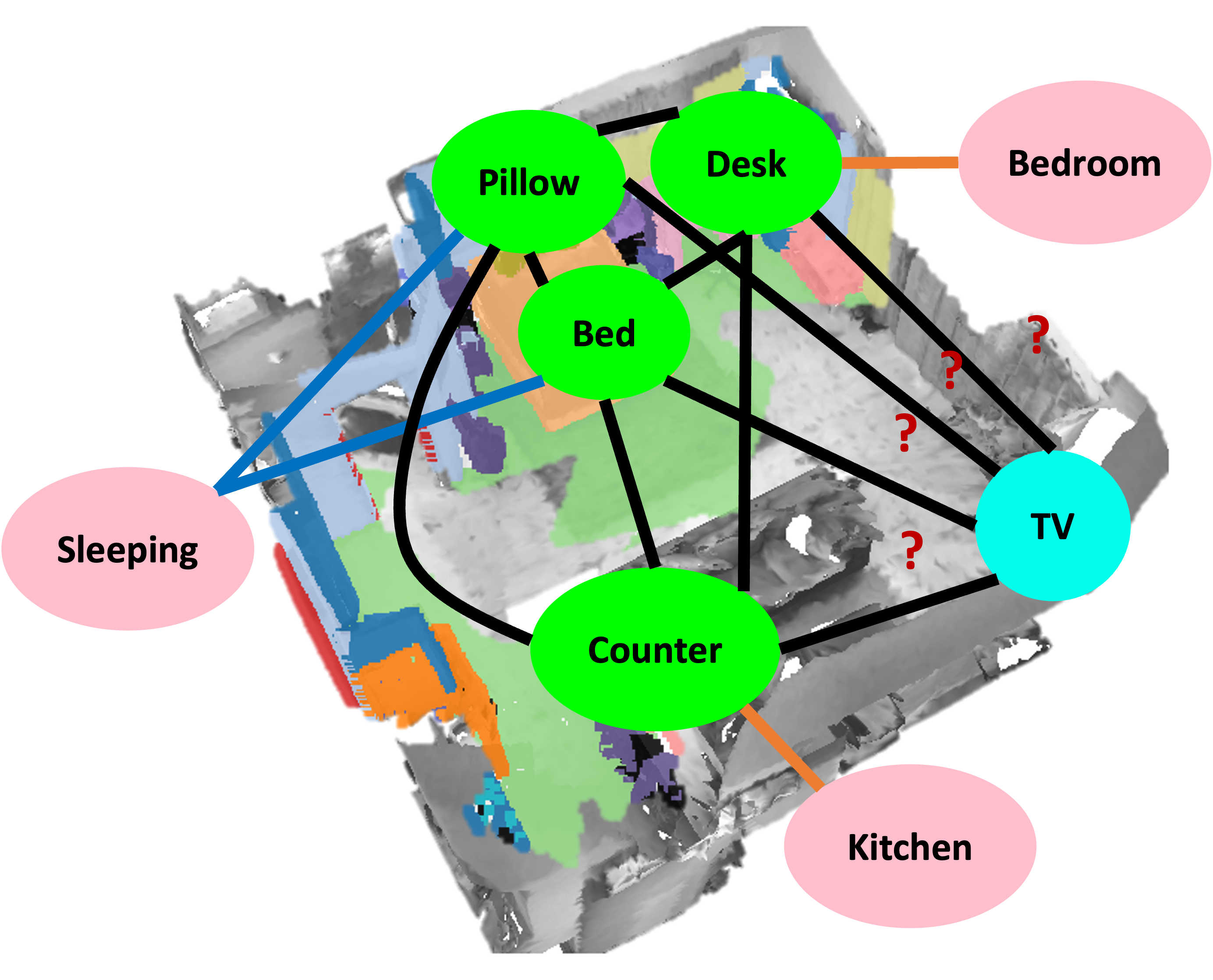}
	\vspace{-0.5cm}
	\caption{Given a set of objects ({\textbf{green}} nodes) in a partially known scene that is semantically coloured, we aim to estimate the position of a target object (the {\textbf{cyan}} node) in the unexplored (grey) area. We address this localisation problem with a novel scene graph representation dubbed \newgraph, that contains both the spatial knowledge extracted from the reconstructed scene, i.e. the proximity (\textbf{black} edges), and the commonsense knowledge represented by a set of relevant concepts ({ \textbf{pink}} nodes) connected by relationships, e.g. {\it UsedFor} ({\textbf{blue}} edges) and {\it AtLocation} ({\textbf{orange}} edges).} 
	\label{fig:teaser}
\end{figure}

\IEEEraisesectionheading{\section{Introduction}
\label{sec:introduction}}
Localising an unobserved object given only a partial observation of a scene, as shown in Fig.~\ref{fig:teaser}, is a fundamental task in many automation applications such as object search with embodied agents~\cite{batra2020objectnav}, layout generation for interior layout design~\cite{luo2020end}, and for assisting visually impaired people in finding everyday items. Humans can perform such a task with quite trivial efforts based on both our past experience and the fact that there exists some commonsense in terms of object arrangement patterns within specific scenarios. For example, when we arrange objects in a house, we often place the television in front of a sofa in the living room, and put the nightstand beside the bed in the bedroom.

In this work, we present a novel solution for the localisation of objects in partially observed 3D scenes. Our method is able to infer the position of an object in the unobserved part of the room by leveraging the commonsense knowledge together with the geometric arrangement of objects in the visible part. We demonstrate that the \emph{commonsense} knowledge that indicates \emph{how objects are used}, e.g., a chair is used for sitting, and \emph{where they are typically located}, e.g., a chair is often located in a kitchen, can be exploited to improve the accuracy of the localisation.

As shown in Fig.~\ref{fig:teaser}, we propose to model the object arrangement and commonsense information as a heterogeneous graph called Directed Spatial Commonsense Graph~(\newgraph). Firstly, the nodes representing the known objects in the partially observed scene construct a Directed Spatial Graph ({\bf \newsg}), which is fully connected. The edges between the nodes are called proximity edges, representing the relative position between a pair of objects. Then, the \newsg\ is further expanded into the \newgraph\ by adding and connecting nodes that represent concepts through relevant commonsense relationships extracted from ConceptNet \cite{speer2018conceptnet}. The object to locate, i.e. the target, is a node in the graph, connected to the other known object nodes with proximity edges, where the respective relative positions are treated as unknowns. Our proposed solution for object localisation exploits a Graph Neural Network (GNN) that can efficiently learn the representations of the nodes and edges of \newgraph. To facilitate the graph representation learning, we introduce a novel attention module with a high expressive power that encourages the sparsity of attention weights, while being computationally efficient. Our network predicts the relative positions between the target object node and each known object node. It then converts the relative positions into absolute ones which agree on a single final position of the target object. 

Differently from the previous work that localises object with Spatial Commonsense Graph (SCG-OL)~\cite{Giuliari2022Spatial}, our new \newgraph\ represents the proximity edges with directional relative positions, instead of the undirected relative distances. This allows us to regress and estimate the target position in an end-to-end trainable manner, without requiring a non-differentiable multilateration procedure for localising the target object, thus contributing to a more effective loss calculation and training procedure. We validate our proposed method with extensive experiments and demonstrate that it achieves an increase of 5.9\% in terms of localisation success rate, compared to the previous state-of-the-art method SCG-OL, with a 8x speed-up in both training and inference. Furthermore, the proposed solution is able to generalise to the 3D domain, reaching 25\% in term of localisation success rate compared to the 5\% of SCG-OL.

Our main contributions can be summarised as below:
\begin{itemize}[noitemsep,nolistsep]
\item We introduce the \textbf{Directed Spatial Commonsense Graph} (\textbf{\newgraph}), an heterogeneous scene graph representation that integrates both the spatial information of the partially observed scene and the commonsense knowledge that is relevant to the observed objects.

\item We propose {\bf \newgraph\ Object Localiser}, a GNN-based solution that uses the \newgraph\ for the localisation of objects in the unobserved part of the scene.

\item We propose a novel attention module during message passing which encourages weight sparsity, and we demonstrate that our novel design can contribute to a higher object localisation success rate compared to existing attention modules. 

\item We present extensive evaluation of our proposed solution, as well as an in-depth analysis of the internal working of the proposed solution to explain the utility of commonsense reasoning.
\end{itemize}

This work is an extended version of~\cite{Giuliari2022Spatial}. The novel contributions with respect to our previous findings are: i) We propose \newgraph\ that describes the spatial information on the proximity edges via directional relative positions, which improves the previous formulation of the SCG that represents the proximity edges as distances. The new formulation removes the need for a non-differentiable multilateration phase when computing the position of the target object as in~\cite{Giuliari2022Spatial}, leading to a 8x speed-up for training and a much better generalisation capability when applied in the 3D domain. ii) We propose a new GNN-based solution with \newgraph\ to address the object localisation task, achieving a 5.9\% increase in the localisation success rate.
iii) We introduce a novel attention module which improves both the efficiency and accuracy in terms of object localisation.

\begin{figure*}[t!]
	\centering
	\scalebox{1}[1]{\includegraphics[width=\linewidth]{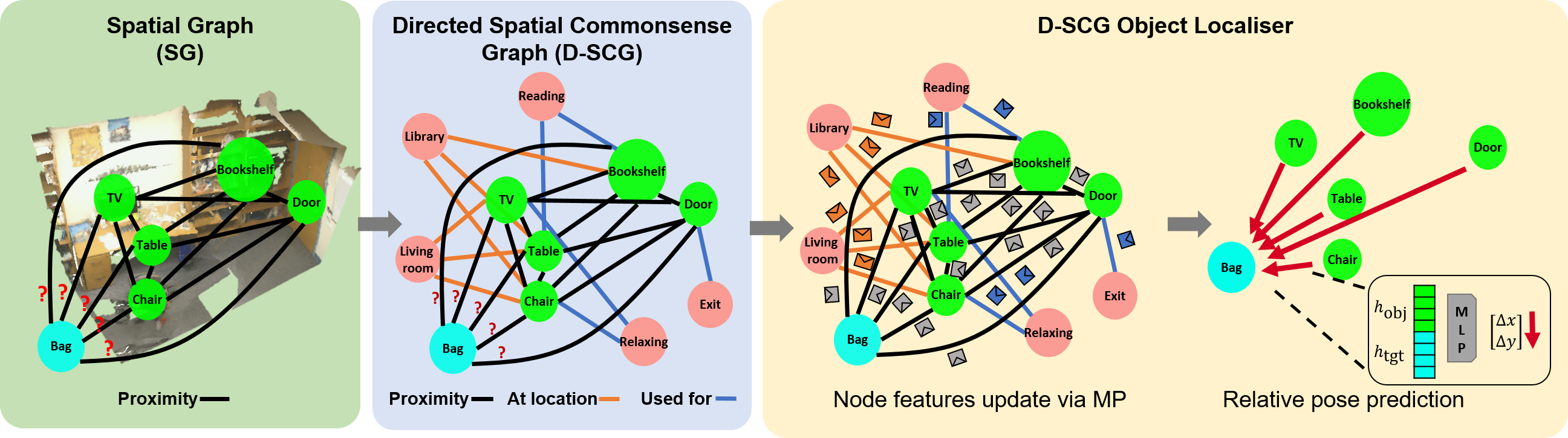}}
	\caption{General overview of our proposed approach. First, we construct the Directed Spatial Commonsense Graph (\newgraph) from the known scene by enriching the directed scene graph with concept nodes and relationships, resulting in edges of three types: {\it UsedFor} ({\textbf{Blue}} edges), {\it AtLocation} ({\textbf{Orange}} edges) and Proximity (edges). The \newgraph~is then fed into our \newmnamefull~that first performs message passing with attention to update the node features taking into consideration the heterogeneous edges, then concatenates the node features of the target node and one of the scene object nodes (at a time) and passes it through an MLP to predict the relative position. The final position is given by aggregating all the predicted relative positions via mean pooling.
	}
	
	\label{fig:architecture}
\end{figure*}

\section{Related Work}

\label{sec:sota}

We will cover prior works that are related to the use of graphs as structures for modelling scenes and performing inference, and the use of commonsense reasoning in neural network architectures.

\noindent\textbf{Scene graph modelling and inference.}
Scene graphs allow high-level description of a scene by its content. They were initially introduced to describe the objects in an image and their relations. In most applications, the nodes in the scene graph indicate the objects in the image, and the edges define the relationship between these objects, which can be spatial~\cite{liu2018structure} or semantical~\cite{Johnson2015RetrievalGraph}. Scene graphs are useful in many applications where using a high-level, abstract representation of the scene is better than directly working in the pixel space. Such applications include image retrieval~\cite{Johnson2015RetrievalGraph, Schuster2015GeneratingSP}, image captioning~\cite{xu2019scene, yang2019auto, Gu_2019_ICCV} and visual question answering~\cite{Shi_2019_CVPR, lee2019visual}. Scene graphs have also been shown to be useful in improving classical tasks like object detection by allowing reasoning on contextual cues from neighbouring objects~\cite{liu2018structure}.

Recently, their usage has been extended to 3D, providing an efficient solution for 3D scene description. The scene graphs for 3D applications can vary from simple structures, where the nodes define the objects in the scene, and the edges define the spatial relationship between them~\cite{gay2018visual, Wald20203Dssg, Wu_2021_Scenegraphfusion}, to more complex, hierarchical structures. Armeni et al.~\cite{armeni20193d} propose a hierarchical scene graph for large-scale environments that can encode information at different "levels", where each level provides a more abstract representation. The first level records the data retrieved from the individual cameras, such as images and camera positions. The second level provides information regarding the objects in the environment, the third information about the rooms, and finally, the last layer stores information regarding the buildings. Such representation is ideal for large environments, but it is needlessly complex for most applications where the environment is composed of a few rooms, so more straightforward representations are typically used. 
3D scene graphs are used for layout completion, scene synthesis and robot navigation. The work of Zhou et al.~\cite{Zhou2019SceneGraphNetNM} uses a GNN in combination with a 3D scene graph to enrich indoor rooms with new objects that match their surroundings. In ~\cite{wang19PlanIT} the layout of the scene is encoded using a relation graph with objects as nodes and spatial/semantic relationships between objects as edges. The relation graph is then used to train a generative model that produces novel relation graphs and thus new layouts. For robot navigation, the scene graph is used to encode the environment efficiently. In~\cite{ravichandran2022Graphnav} the scene graph is used to encode the scene’s geometry, topology, and semantic information. The scene graph is then mapped to the robot's control space used for navigation via a learned policy.
In this work, we propose \newgraph\ to address the object localisation task in unknown areas. The spatial relationship between objects is described with relative positions instead of distances, resulting in more a accurate localisation. The proposed scene graph is used to encode both the geometrical information regarding the arrangement of the objects in the observed part of the room and commonsense attributes that define what they are \emph{commonly} used for or where they are \emph{commonly} located.\newline

\noindent\textbf{Commonsense Knowledge in Neural Networks.}
Commonsense reasoning refers to the high-level reasoning that humans employ when solving tasks. In particular, it is our ability to use prior information gained in our lifetime and use it for a new task. 
While modelling human-level commonsense is something that we are still far away from achieving, much work has been done in this direction in recent years. A fundamental requirement is to have a way to provide "known prior information". This is typically achieved using knowledge bases that are considered to contain some kind of axiomatic truths regarding our world. Examples of such knowledge bases are WikiData~\cite{vrandevcic2014wikidata} and ConceptNet~\cite{speer2018conceptnet}. 

In the field of Natural Language Processing (NLP), the work presented in \cite{faldu2021kibert} makes use of ConceptNet to create richer contextualised sentence embeddings with the BERT architecture~\cite{devlin2019bert}. In~\cite{bao2014knowledge}, the authors utilise the knowledge graph Freebase to enrich textual representations for a question answering system. In computer vision, a few works~\cite{li2017incorporating, NARAYANAN2021104328} have exploited an external knowledge base for Visual Question Answering (VQA) as it helps the network to reason beyond the image contents. In the scene graph generation task, the ConceptNet~\cite{speer2018conceptnet} knowledge graph has also been exploited to refine object and phrase features to improve the generalisation of the model~\cite{gu2019scene}. 
In this work, we incorporate commonsense information from ConceptNet into the spatial scene graph to improve the object localisation performance when only a partial scene is observed.

\section{Directed Spatial Commonsense Graph}
\label{sec:graph_modelling}
Our scene representation has the objective to embed commonsense knowledge into a geometric scene graph extracted from a partial 3D scan of an area. 
As shown in Fig.~\ref{fig:architecture}, we construct the \newgraph~with nodes that are: \textit{i)} object nodes that include all the observed objects in the partial scene and any unseen target object to be localised; \textit{ii)} concept nodes that are retrieved from ConceptNet~\cite{speer2018conceptnet}. 

Each \newgraph~is constructed on top of a directed Spatial Graph (\newsg), a fully directed graph with all object nodes. Each object node is further connected to a set of concept nodes via some semantic relationships available in the knowledge base. This renders the edges of \newgraph~heterogeneous, separating the spatial interactions from the "commonsense". 
In practice, the edges of our proposed graph structure are of three types:
\begin{itemize}[noitemsep,nolistsep]
    \item \emph{Proximity}, represented by the relative position vector, indicating both the distance and direction, between \emph{all} object nodes given the partial 3D scan. This is different from our previous work \cite{Giuliari2022Spatial} where \emph{Proximity} is represented by the relative distance between \emph{all} the object nodes of the partial scene;
    \item \emph{AtLocation}, retrieved from ConceptNet, indicating in what environment the objects are often located in; 
    \item \emph{UsedFor}, retrieved from ConceptNet, describing common use-cases of the objects.
\end{itemize}

The proximity edges connect all the objects nodes of the \newgraph~in a directed and complete manner, while the semantic \emph{AtLocation} and \emph{UsedFor} edges connect each object node with its related concept nodes that are queried from ConceptNet (e.g. \textit{bed AtLocation apartment} or \textit{bed UsedFor resting}). The two semantic edge types provide useful hints on how objects can be clustered in the physical space, thus benefitting the position inference of indoor objects.

We formulate \newgraph~as a directed graph that is composed by a set of nodes $\mathcal{H}=\{ h_i| \: i\in(0,N]\}$,
where $N = N_o+N_c$ is the total number of nodes. $N_o$ is the number of the object nodes and $N_c$ the number of the concept nodes, where each node is represented by a feature vector $h_i\in\mathbb{R}^{300}$ from ConceptNet~\cite{speer2017numberbatch}.
The edges are defined by the set  $\mathcal{E}=\{ e_{i,j}| \: {i\in(0,N], j\in\mathcal{N}_i}]$, where $e_{i,j}$ is the edge between node $i$ and node $j$ and $\mathcal{N}_i$ is set of neighbouring nodes of $i$.

We represent each edge with a 6-dimensional feature vector, i.e., $ e_{i,j} \in\mathbb{R}^{6}$, whose first three elements indicate the edge type in a one-hot manner, the fourth element indicates whether a proximity relation involves the target node, while the last two elements indicate the relative position $d_{i,j} =[\Delta x_{i,j},\Delta y_{i,j}]$ between node $i$ and node $j$, in Cartesian coordinates such that $\Delta x_{i,j} = x_j - x_i$ and $\Delta y_{i,j} = y_j - y_i$. This definition is different from the \oldgraph\ in \cite{Giuliari2022Spatial}, where edges were represented by 4-dimensional vectors that represented the one-hot encoded edge class and only the distance between the objects that were connected by the edge. In our new graph formulation, we are able to achieve a more detailed spatial reasoning and to train in an end-to-end manner, which contributes to the performance improvement in both object localisation and computational efficiency as shown in Sec.~\ref{sec:exp_val}. 
As the relative positions are only measurable among object nodes in the observed part of the 3D scan, we initialise the relative positions to $[0,0]$ when the edges are \textit{AtLocation}, \textit{UsedFor}, or \textit{Proximity} edges involving the unknown target object node. 

Note that we focus on localising the target object on the XY plane since the target's positions vary little along the Z axis, as indicated in the benchmark's statistics, shown in~\cite{Giuliari2022Spatial}. Nevertheless, our method can easily be extend to perform predictions in 3D space by predicting the relative 3D positions $d_{i,j} =[\Delta x_{i,j},\Delta y_{i,j}, \Delta z_{i,j}]$ between nodes $i$ and $j$, in Cartesian coordinates. We provide experimental results on 3D localisation in Sec.~\ref{sec:exp_ablation}.

\section{End-to-End \newmnamefull}
\label{sec:method}
We propose an end-to-end solution to address the task of localising the arbitrary unobserved target object using the \newgraph. The model first predicts the relative positions of the unseen target object w.r.t. the objects in the partially known scene. Then the relative positions are converted in absolute coordinates and mean pooling is applied to estimate the final position. This approach is fully differentiable and requires no additional localisation module based on circular triangulation to predict the position of the target object as in~\cite{Giuliari2022Spatial}, thus improving the network's efficiency.

\subsection{Model}
\begin{figure}
    \centering
    \includegraphics[width=\linewidth]{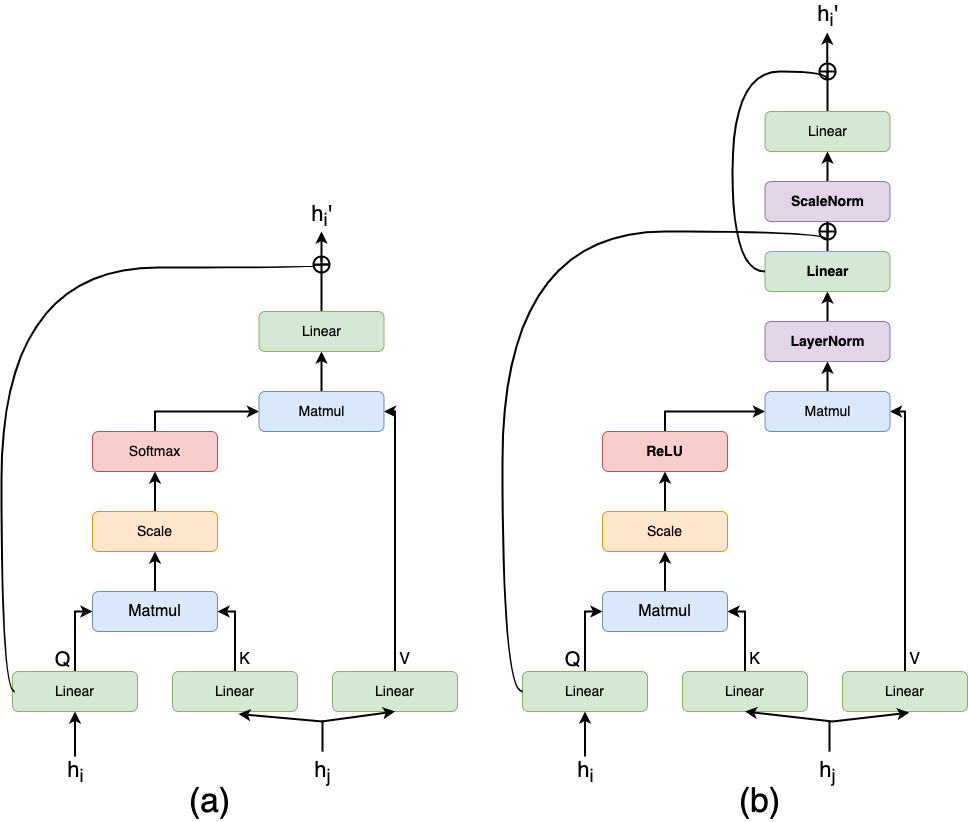}
    \caption{Overview of the differences between the attention mechanism of~\cite{shi-graphtransformer} (a), used in \oldgraph~\cite{Giuliari2022Spatial}, and the one proposed in this paper (b). The new attention mechanism contains more parameters, thus producing more expressive representations, and learns sparse weights with reduced training and inference time thanks to the ReLU activation function. Moreover, we utilise two different normalisation layers to stabilise the network's training. The changes from (a) to (b) are highlighted in \textbf{bold}.}
    \label{fig:attn_mech}
\end{figure}

To predict the relative position of the unseen target node w.r.t. the visible scene objects, we make use of a stacked GNN architecture. Our proposed GNN relies on an novel attentional message passing mechanism, that modifies the Graph Transformer~\cite{shi-graphtransformer} by implementing a sparse attention mechanism with Rectified Linear Unit (ReLU) Activation and improving the training scale by using different normalisation layers. The node embeddings are updated iteratively by utilising the heterogeneous information of the edge type, to allow effective fusion between the commonsense knowledge and the metric measurements. We highlight the main differences between the attention mechanism in~\cite{shi-graphtransformer} and ours in Fig.~\ref{fig:attn_mech}

The input to the network is a set of node features $\mathcal{H}$ and the output is a new set of node features $\mathcal{H}^{'}=\{ h_i'| \: i\in(0,N]\}$, with $ h_i'\in\mathbb{R}^{300}$. Each node $i$ in the graph is updated by aggregating the features of its neighbouring nodes $\mathcal{N}_i$ via four rounds of message passing. The resulting $ h_i'$ forms a \emph{contextual} representation of its neighbourhood.

At each round of the message passing, we learn an attention coefficient $\alpha_{i,j}$ between each pair of connected nodes using a graph-based and rectified version of the scaled dot-product attention mechanism~\cite{zhang-etal-2021-sparse}, conditioned on the node and edge features.
Our GNN can learn sparse attention weights due to the usage of the activation function ReLU, thus allowing for the understanding of arbitrary relationships between the different node types.

The network starts by performing an affine transformation of the relevant node and edge features to calculate the corresponding query, key, value and edge vector that will be used to compute the attention weights:
\begin{align}
     q_i & = {W_q} h_i + b_q, \label{eq:1} \\
     k_j & = {W_k} h_j + b_k, \label{eq:2} \\
     v_j & = {W_v} h_j + b_v, \label{eq:3} \\
     e_{ij} & = {W_e} e_{ij} + b_e, \label{eq:4}
\end{align}
where $W$ and $b$ represent respectively the learnable weight matrices and bias vectors for each transformation. 

The network then calculates the attention weight $\alpha_{i,j}$ between two nodes $i$ and $j$ as:
\begin{equation}
     \alpha_{i,j}  = ReLU(\frac{\langle q_i, k_j + e_{ij} \rangle}{\sqrt{d}} )
\end{equation}
where $\sqrt d$ is a scaling term equal to the square root of the dimension of the projected features $k_j$. Inspired by Zhang et al~\cite{zhang-etal-2021-sparse}, we use ReLU instead of the classical softmax to activate the learned attention weights, as using softmax involves aggregating the scores for all the edges connected to each node, which implies many operations for large graphs and slows down the training. Additionally, by using ReLU, we allow for sparsity in the attention weights, which helps when analysing how the network prioritises the exchange of information. As the attention weights that are calculated using ReLU are not limited to the range $(0,1)$, we use Layer Normalisation~\cite{ba_layernorm} when calculating the updated node features $h_i'$, followed by a gated residual connection that prevents the node features from converging into indistinguishable features~\cite{shi-graphtransformer}:
\begin{align}
     h_i' = LayerNorm(h_i + \langle \alpha_{i,j}, v_j + e_{ij} \rangle)   \label{eq:6} \\
     \beta_i = Sigmoid( W_g[ h_i'; W_r h_i +b_r; h_i' - (W_r h_i - b_r)])\\
     h_i' = (1-\beta_i){h}_i' + \beta_i ({W_r} h_i + b_{r}), \label{eq:7}
\end{align}
where $[;]$ represents the concatenation operation, $LayerNorm$ is the Layer Normalisation, $\beta_i$ is a learnable parameter that guides the gated residual connection, and $W$ and $b$ are the learnable weight matrices and bias vector for the respective linear transformations. 

Differently from typical graph attentional networks, these features are then re-projected and re-normalised in a similar fashion to the original Transformer model, a practice that has been empirically shown to stabilise and improve the training of self-attentive neural networks. 
\begin{gather}
     h_i' = ScaleNorm({W_o} h_i' + b_o), \label{eq:8}
\end{gather}
This step further increases the number of learnable parameters of our GNN, allowing for better scaling and more expressive representations, while not sacrificing efficiency thanks to the sparse attention mechanism (described previously) and the Scale Normalisation introduced in~\cite{nguyen-salazar-2019-transformers}.

We combine all these operations in a module, and use it for a total of four message passing rounds. Finally, we obtain the set of final node embeddings $\mathcal{H}^{*}=\{{h}_i^*| \: i\in(0,N]\}$, with $ h_i^*= [h_i; {h'}_i]$. In this way, the final representation of each node contains both the original object embedding and the aggregated embedding of its context in the scene. Finally, we concatenate the features of the two nodes $ h_{i,t}^*=[{h}_i^*; {h}_t^*]$, and predict the relative position $\hat{d}_{i,t}$ between the target object node $t$ and the observed object node $i$ via linear projection.
To obtain the final position, we first convert the relative positions $\hat{d}_{i,t}$ in absolute coordinates by summing to them the positions $p_{i} = [x_i,y_i]$ of the observed object nodes and then take the mean of the absolute positions as our predicted position $\hat{p_t}$.

\subsection{Loss}
We train our network with a strategy which considers that multiple instances of the searched object can exist in the unobserved part of the scene. Therefore, only the instance closest to the prediction is accounted when calculating the loss. By doing this, the network learns to correctly predict a specific position instead of a point that minimises the distance w.r.t. all the instances. For the loss, we minimise the squared L2 distance between the predicted position $\hat{p_t}$ and the ground-truth position of the target position $p_t$ as follows:
\begin{equation}
    \mathcal{L}_{\text{2}}(\hat{p_t},p_t)= \Vert \hat{p_t} - p_t \Vert_2^2.
    \label{loss:mse}
\end{equation}

\section{Experiments}
\label{sec:exp}
We evaluate our proposed method on a dataset of partially reconstructed indoor scenes. In the following sections, we first give some relevant details on the partial scene dataset in Sec. \ref{sec:exp:dataset}. Then we present comparisons of our proposed method against the state-of-the-art methods, accompanied by the implementation details, evaluation metrics and discussions in Sec. \ref{sec:exp_val}. Finally, we show different ablation studies in Sec. \ref{sec:exp_ablation} to prove our main design choices and to demonstrate some interesting aspects of our method, including how the proposed attention evolves over message passing and the extension towards 3D object localisation.

\subsection{Dataset}
\label{sec:exp:dataset}
Our training and evaluation is based on the partial 3D scenes dataset~\cite{Giuliari2022Spatial}. The dataset is built using data from ScanNet~\cite{dai2017scannet} which contains RGB-D sequences taken at a regular frequency with a RGB-D camera, providing the camera position corresponding to each captured image, as well as the point-level annotations, i.e., class and instance id, for the complete Point Cloud Data (PCD) of each reconstructed scene.

\begin{figure*}[t]
    \centering
    \includegraphics[width=0.8\linewidth]{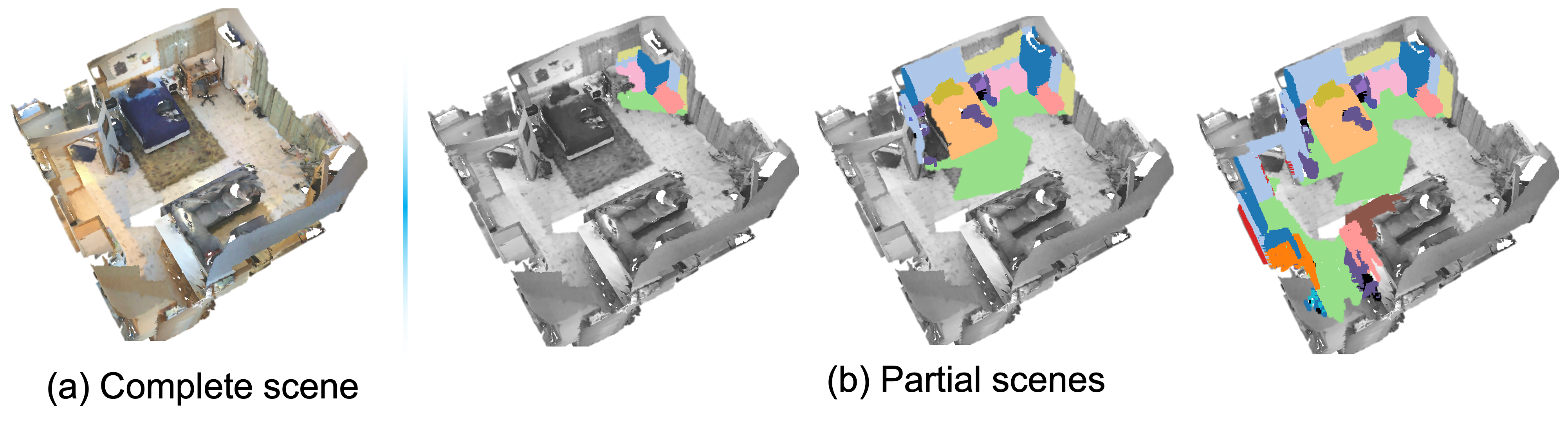}
    \vspace{-0.3cm}
    \caption{The dataset with (a) the complete scene from the ScanNet~\cite{dai2017scannet} dataset, and (b) the extracted partial scenes~\cite{Giuliari2022Spatial}. The observed part of the scene is coloured based on the object semantics, while the unexplored part of scene is coloured in grey.}
    \label{fig:dataset_sample}
\end{figure*}

As the original acquisition frequency in ScanNet is very high (30Hz), the partial scene dataset only uses a subset provided in the ScanNet benchmark\footnote{http://kaldir.vc.in.tum.de/scannet\_benchmark} with a frequency of about $1/100$ of the initial one. Each full RGB-D sequence of each scene is divided into smaller sub-sequences to reconstruct the partial scenes, with varying length to reflect different levels of completeness of the reconstructed scenes (see Fig.~\ref{fig:dataset_sample} for an example). For each sub-sequence, the RGB-D information is integrated with the camera intrinsic and extrinsic parameters to reconstruct the PCD at the resolution of 5cm using Open3D~\cite{open3d}. The annotation for each point in the partial PCD is obtained by looking for the corresponding closest point in the complete PCD scene provided by ScanNet.  

We extract the corresponding \newsg, i.e., the graph with only proximity edges, for each partially reconstructed scene and its object nodes. The nodes of the graph contain the object information: its \emph{position}, defined as the centre of the bounding box containing the object and the \emph{object class}. The proximity edge connects two object nodes and contains the relative position of the second object with respective to the first. We consider the position of each scene object as a 2D point $(x,y)$ on the ground plane as most objects in the indoor scenes of ScanNet are located at a similar elevation. Each node is marked as \emph{observed} if it represents an object in the partially known scene; or as \emph{unseen} if it represents the target object in the unknown part of the scene.

\begin{figure}[t]
          \centering
          \includegraphics[width=1\linewidth]{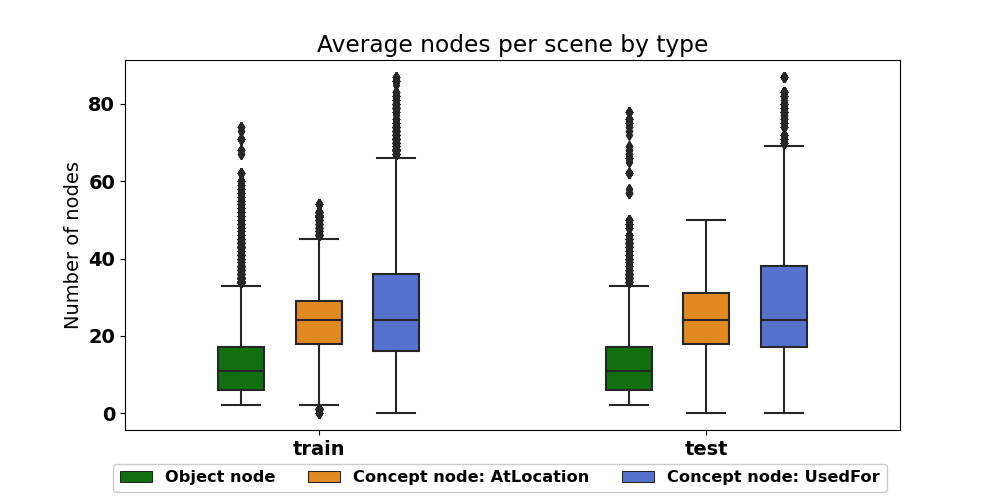}
          \caption{Average number of different types of nodes in the \newgraph~for both train and test splits of the dataset. The outliers in the boxplots are introduced by uncommon room types with a large amount of objects, e.g. libraries with several books.} 
          \label{fig:dataset_stats}
\end{figure}

\begin{figure*}
    \centering
    \subfloat[][Scene]{\centering\includegraphics[width=.40\linewidth]{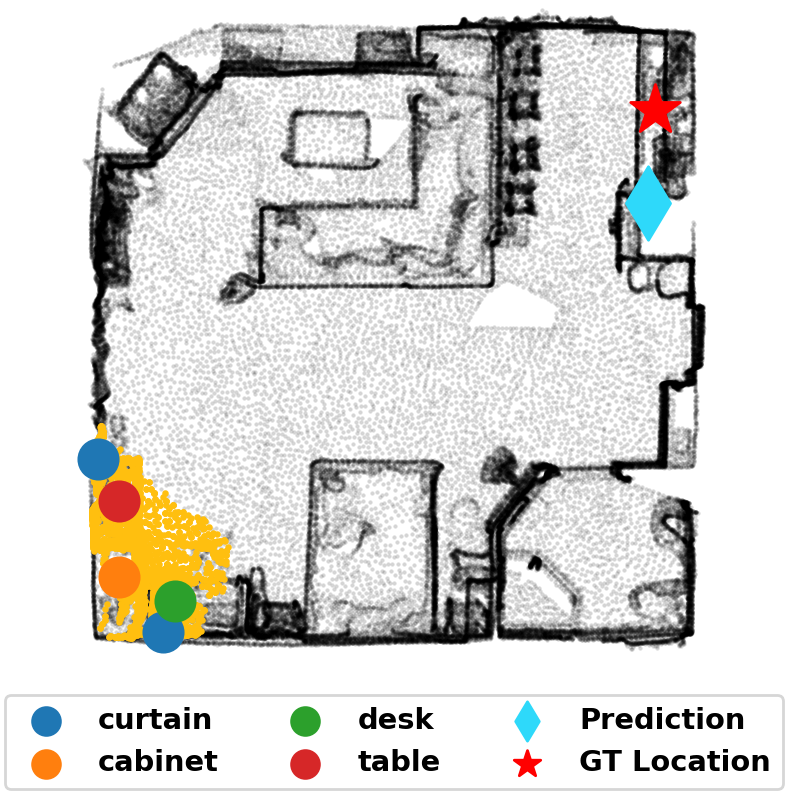}}
    \hfill
    \subfloat[][Directed Spatial Commonsense Graph]{\centering\includegraphics[width=.55\linewidth]{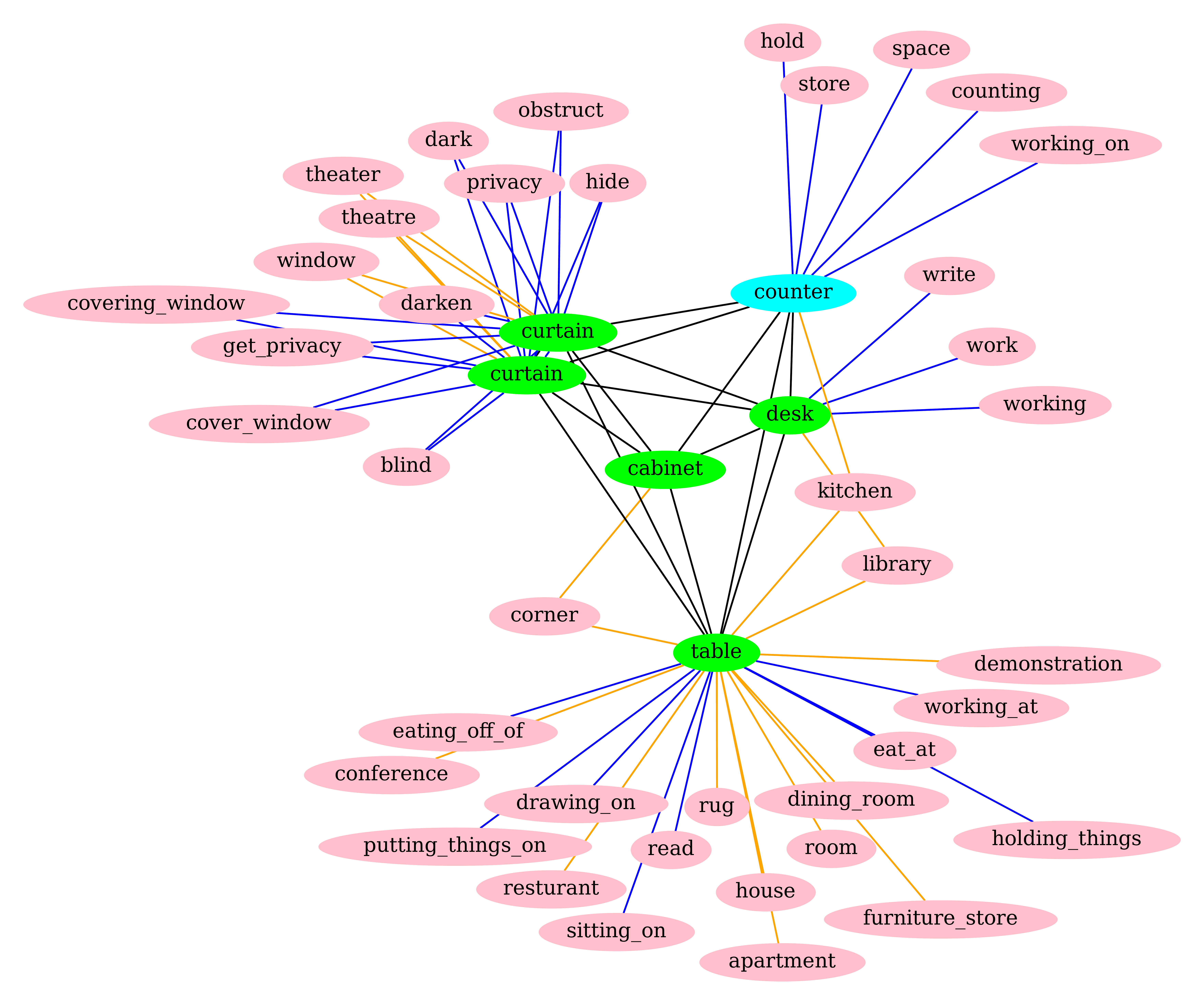}}
    \caption{Example of a Partial Scene and its generated \newgraph. The target object is represented by the {cyan node}, the scene objects are the {green nodes}, and the concept nodes have a {pink background}. The colour of the edge distinguishes the relationship type: {orange ones} are \emph{AtLocation} edges, {blue ones} are \emph{UsedFor} edges, and \textbf{{black ones}} are \emph{Proximity} edges.}
    \label{fig:SCG_example}
\end{figure*}

On top of \newsg, each \newgraph\ is constructed by adding two semantic relationships \emph{AtLocation} and \emph{UsedFor}, as well as the concept nodes that are linked to the scene nodes by these relationships. The concepts are extracted from ConceptNet by querying each scene object using the two semantic relationships. The query returns a set of related concepts together with their corresponding weight $w$, which indicates how ``safe and credible'' each related concept is to the query. We include a concept to the \newgraph\ only when it has a weight $w > 1$. Fig.~\ref{fig:SCG_example} shows an example of a scene and the extracted \newgraph.
Fig.~\ref{fig:dataset_stats} shows the average number of nodes linked by different types of edges in the \newgraph. On average, each \newgraph\ contains about 5 times more the concept nodes than the object nodes in the \newsg, demonstrating that a rich commonsense knowledge is included in \newgraph.

Finally, we follow the same training/validation/test split as in \cite{Giuliari2022Spatial}, %
using $19461$ partial scenes for training and validation, and $5435$ partial scenes for testing, with each partial scene having its corresponding \newsg~and \newgraph. 

\subsection{Experimental Comparisons}
\label{sec:exp_val}
We validate \newmname~by comparing its performance against a set of baselines, a method for layout prediction adapted for the localisation task~\cite{gupta2021Layout} and the state-of-the-art approach SCG-OL for object localisation~\cite{Giuliari2022Spatial}.
The baselines and SCG-OL follow the two-staged pipeline, where they first predict the pairwise distances and then estimate the position by using a localisation module, which minimises these pairwise distances (circular intersection). 
We summarise  all the approaches implemented for evaluation below.
\begin{itemize}[noitemsep,nolistsep]
    \item \textbf{Statistics-based baselines} use the statistics of the training set, i.e., the \textit{mean}, \textit{mode}, and \textit{median} values of the pairwise distances between the target object and the scene objects, as the predicted distance.
    
    \item \textbf{MLP} learns to predict pairwise distances between the target object and every other observed object in the scene without considering the spatial or semantic context. The input to this model is a pair of the target object and the observed object with each object represented by a one-hot vector indicating the class, passed to an MLP that predicts pairwise distances. %
    
    \item \textbf{MLP with Commonsense} learns to predict the pairwise distance between the target object and every other observed object in the scene without considering the spatial context. We first use GCN to propagate the conceptnet information to object nodes, then the features are passed to a MLP that predicts pairwise distances. 
    
    \item \textbf{LayoutTransformer}~\cite{gupta2021Layout} uses the transformer's self-attention to generate the 2D/3D layout in an auto-regressive manner. We describe the observed objects as a sequence of elements as in \cite{gupta2021Layout}, where each element contains the object class and the position $(x, y)$. We then feed the class of the target object to generate its corresponding position $(x, y)$. For a fair comparison, we retrain the model on our training set.
    
    \item \textbf{\mname} \cite{Giuliari2022Spatial} exploits \oldgraph~that contains the proximity edges that describe the relative distances. The localisation method is a two-stage approach which first predicts the pairwise distances via a stacked GNN, and then passes these to a Localisation module to obtain the final position.
    
    \item \textbf{\textbf{\newmname~ w\textbackslash o Commonsense}} 
    is a variant of our approach to test the capability of the method when it is used without commonsense knowledge. The input is the \newsg, which is composed only by the object nodes and proximity edges. The initial node features are not pre-trained word embeddings, but are learned during training via an embedding layer. 
    
    \item \newmname~is our proposed method, described in Sec.~\ref{sec:method}, along with a variant that are trained with learnable node embeddings instead of pretrained node embeddings from ConceptNet's Numberbatch.
\end{itemize}
We evaluate the different methods for the localisation on the 2D floor plane, and report additional results for 3D localisation in the ablation studies.

\noindent\textbf{Evaluation Measures.} We evaluate the performance in terms of the successful target object localisation and the relative pairwise distances, as also proposed in~\cite{Giuliari2022Spatial}. 
\begin{itemize}[noitemsep]
    \item \textit{Localisation Success Rate (LSR)} quantifies the localisation performance. LSR is defined as the ratio of the number of successful localisations over the number of tests. A localisation is considered successful if the predicted position of the target object is close to a target instance within a predefined threshold. Unless stated differently, the distance threshold is set to 1m. We consider LSR as the \emph{main} evaluation measure for our task. 
    
    \item \textit{mean Successful Localisation Error (mSLE)} quantifies the localisation error among successful cases. mSLE is the Mean Absolute Error (MAE) between the predicted target position and the ground-truth position among all successful tests.
    
    \item Finally, \textit{mean Predicted Proximity Error (mPPE)} quantifies the performance of the methods that rely on pairwise relative distance prediction, as described above. mPPE is the mean absolute error between the predicted distances and the ground-truth pairwise distances between the target object and the objects in the partially known scene.
\end{itemize}

\noindent\textbf{Implementation Details.} We train our network using the Adafactor optimiser~\cite{pmlr-adafactor} for 200 epochs. We use a total of $4$ message passing layers, a number that is carefully chosen (see details in Sec.~\ref{sec:exp_ablation}). The dimension of the first message passing projection is set to $D=256$ and $2D$ for the remaining rounds. All attention modules use $4$ attention heads. During training, we augment the dataset by applying random rotations to the scene objects to allow for better generalisation. 

\begin{table}[t]
\centering
\caption{Methods comparison for object localisation in partial scenes. 
mPPE: mean Predicted Proximity Error. mSLE: mean Successful Localisation Error. LSR: Localisation Success Rate. SG: Spatial Graph. \oldgraph: Spatial Commonsense Graph. \newsg: Directed Spatial Graph. \newgraph: Directed Spatial Commonsense Graph.
The first part of the table follow the 2-stage approach which first predicts the pairwise distances and the localise the object via multilateration. The last part consists of our method and its variants which directly predict the final position.
}
\resizebox{1\linewidth}{!}{
\begin{tabular}{|c|c|c|c|c|}
\hline
Method &
  Data type &
  mPPE(m)$\downarrow$ &
  mSLE(m)$\downarrow$ & \textbf{LSR} $\uparrow$\\ \hline \hline
  
  Statistics-Mean & Pairwise  & 1.167 &  0.63 & 0.140\\ \hline 
  Statistics-Mode & Pairwise  & 1.471 &  0.63 & 0.149\\ \hline
  Statistics-Median & Pairwise & 1.205 & 0.64 & 0.164\\ \hline
  
  MLP & Pairwise  &  1.165 &  0.62 & 0.143\\ \hline
  MLP w/ Commonsense & Pairwise  &  1.090 &  0.64 & 0.163\\ \hline
  LayoutTransformer~\cite{gupta2021Layout}& List  & - & \textbf{0.59} &0.176\\ \hline
  \mname  - Learned Emb ~\cite{Giuliari2022Spatial} & \oldgraph~\cite{Giuliari2022Spatial} & 0.974 & 0.61 & 0.234\\  \hline
  \mname~\cite{Giuliari2022Spatial} & \oldgraph~\cite{Giuliari2022Spatial} & 0.965 & 0.61 & 0.238\\  \hline \hline

  \newmname~w/o Commonsense & \newsg  & - & 0.59 & 0.265\\  \hline
  \newmname~- Learned Emb & \newgraph & - & 0.57 & 0.273 \\  \hline
  \newmname~ & \newgraph & - & 0.55 & \textbf{0.297} \\\hline

\end{tabular}
}
\label{table:baselines}
\end{table}

\noindent\textbf{Discussion.}
Table~\ref{table:baselines} reports the localisation performance measures in terms of mPPE, LSR, and mSLE, of all compared methods evaluated on the dataset with partially reconstructed scenes. We can initially observe that methods which rely \emph{only} on pairwise inputs, e.g. statistics-based approaches or MLP, lead to worse performance compared to methods that account for other objects present in the observed scene. Nevertheless, introducing semantic reasoning on top of these methods seems to improve the performances, as shown by MLP w/ Commonsense, with an improved LSR of 2\% compared to the standard MLP. LayoutTransformer directly predicts the 2D position of the target object by taking as input the list of all the observed scene objects and using the target class as the last input token. LayoutTransformer can better encode the spatial context and outperforms the statistic-based and MLP baselines. \mname~that uses the \oldgraph~with pairwise distances is able to improve on all metrics w.r.t. the baseline methods, suggesting that a scene-graph based solution with added commonsense knowledge is a more effective way of modelling the problem.

The different versions of our proposed method \newmname~are able to reach the best performance. \newmname~with learned embeddings has a 0.8\% increase in the LSR performance w.r.t. the GNN working only on the \newsg, revealing the usefulness of the concept nodes, with a further increase of 2.4\% when using the ConceptNet's pretrained embeddings, showing that the commonsense information introduced from ConceptNet is useful for the localisation task.

\begin{figure}[t]
	\centering
	\hfill
        \subfloat[][Localisation error\label{fig:abl:A}]{
            \centering
              \includegraphics[width=.45\linewidth]{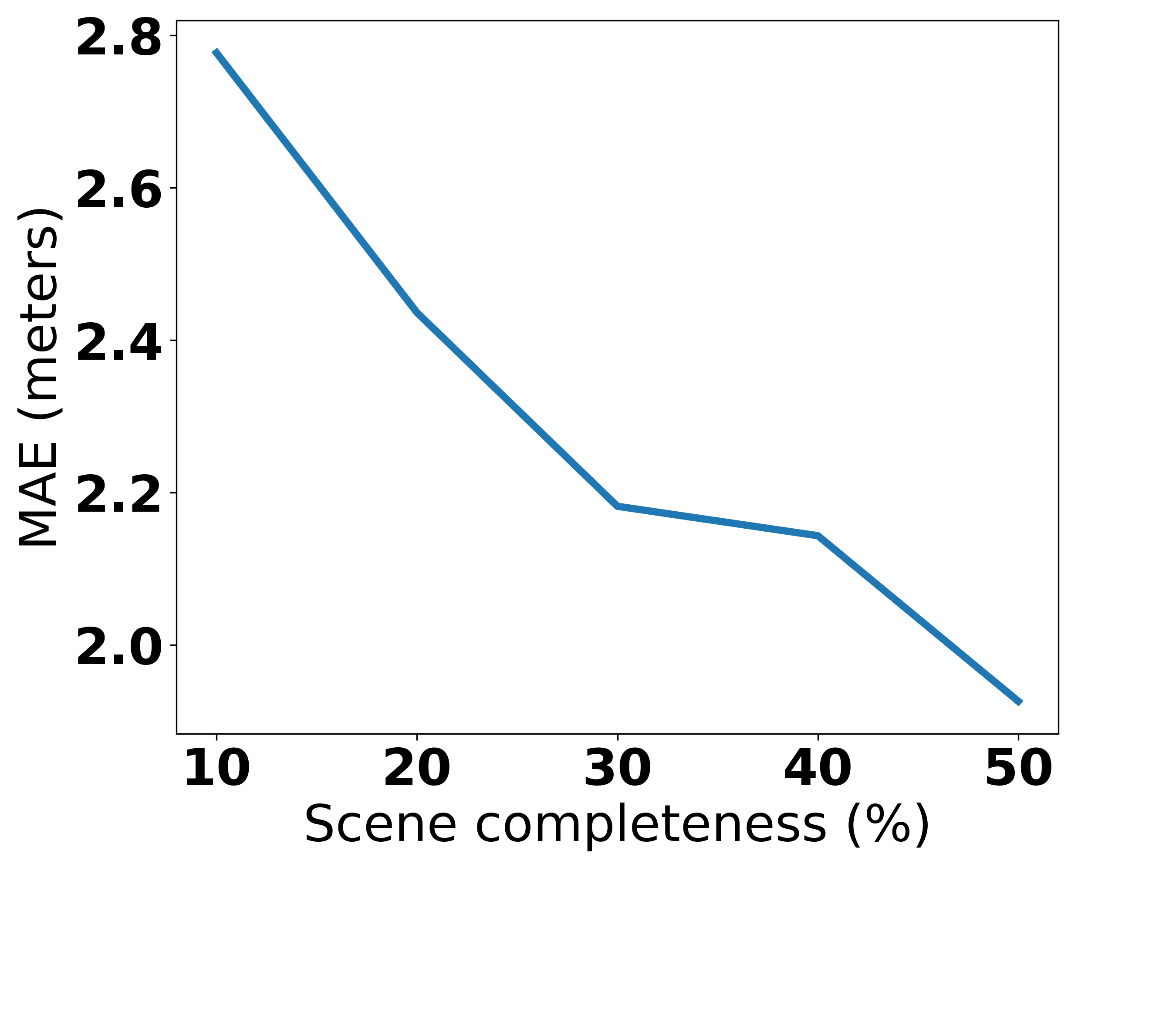}
        }%
        \hfill
        \subfloat[][LSR\label{fig:abl:B}]{
              \centering
              \includegraphics[width=.45\linewidth]{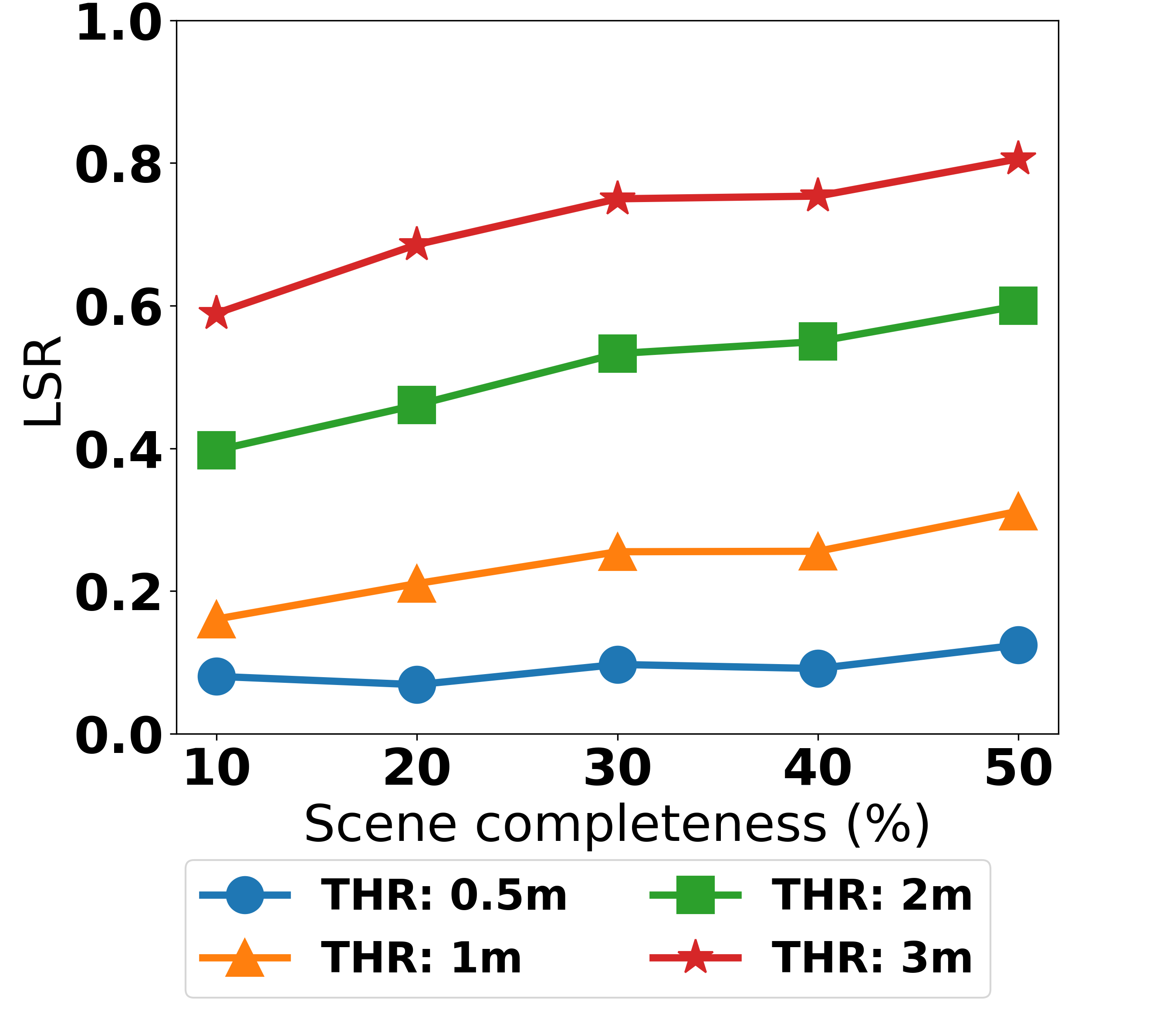}
        }
        \hfill
    \caption{Localisation performance over different levels of scene completeness. (a) The localisation error in terms of MAE between the estimated target position and the ground-truth position. (b) The LSR at different threshold levels.} 
    \label{fig:abl}
\end{figure}

As previously mentioned in this section, we consider the LSR as the primary evaluation metric. It is therefore useful to demonstrate and understand how the completeness (known) level of the scene impacts the localisation performance of \newmname. Fig.~\ref{fig:abl:A} reports the mean absolute error (MAE) between the estimated position and the ground-truth position compared to the scene completeness. Note that the MAE is calculated on all the test cases, including both the successful and failed ones. We use MAE instead of the mSLE as the mSLE is calculated only on successful cases and does not change with the completeness of the scene. As a general trend, our model can predict more accurately the position of the target object with an increasing scene completeness. 
Fig.~\ref{fig:abl:B} presents how the LSR varies as the scene gets more complete. In general, the LSR increases when the localisation error decreases. We report the LSR at four different threshold values, i.e. 0.5m, 1m, 2m, and 3m, where a larger threshold leads to a larger LSR value, as it might be expected.

\begin{figure*}[t]
	\centering
	\includegraphics[width=1\linewidth]{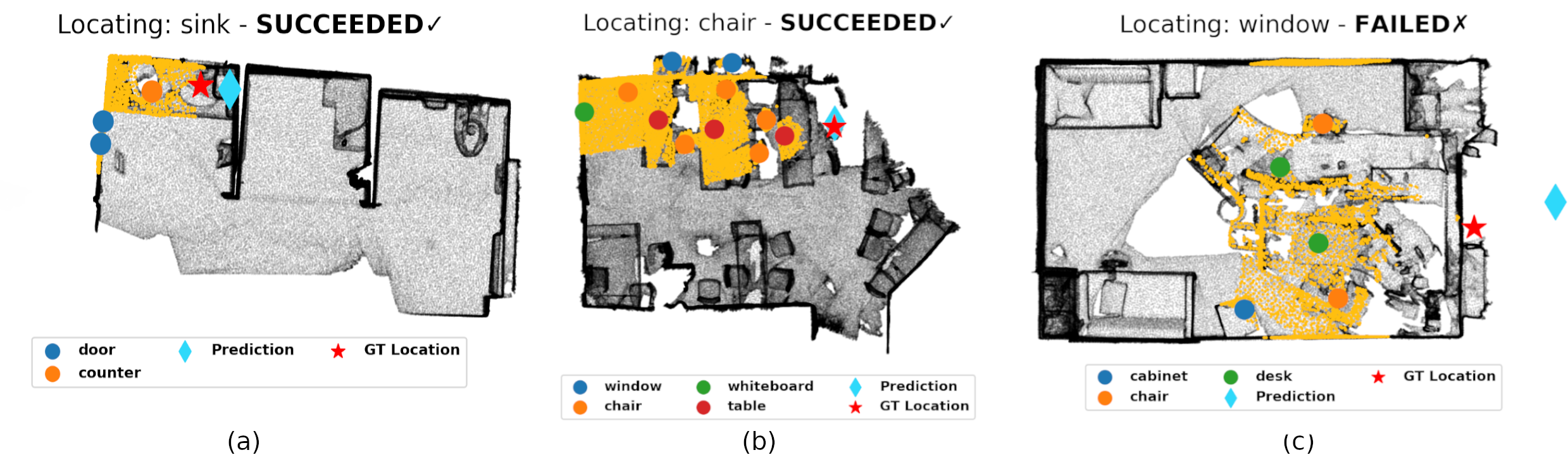}
    \caption{Qualitative results obtained with \newmname. The partially known scene is coloured with a yellow background, while the unknown scene is indicated with grey. The  coloured circles indicate the object nodes present in the \newgraph. The {red star} indicates the GT position of the target object, while the { cyan diamond} indicates the predicted positions. The network is able to correctly predict the position of a sink in (a) and a chair in (b). In the failure case of (c), the network correctly identified the direction of the window but overestimated the distance from the visible objects.}
    \label{fig:test}
\end{figure*}

\noindent{\textbf{Qualitative results.}} 
Fig.~\ref{fig:test} shows the qualitative results obtained using our method \newmname. Fig.~\ref{fig:test}(a) shows that the ``sink'' object class was successfully located near the counter. Similarly in Fig.~\ref{fig:test}(b), the position of the chair (target object) is correctly estimated in a position which is coherent with other instances of chair and tables in the observed part of the room . Interestingly, Fig.~\ref{fig:test}(c) presents a failure case in which the method fails to locate a window in an office setting. In this case the network successfully identify the general direction where the window should be located, but overestimated its concrete placement w.r.t to the visible objects. This error is plausible as the network does not see any objects that can help create an idea of the actual shape of the room.

\noindent{\textbf{Computation efficiency.}} 
There are 20.5M parameters in \newmname, which is 3.4M more than the previous state-of-the-art method SCG-OL \cite{Giuliari2022Spatial} (17.1M). Nevertheless, our proposed \newmname\ takes 13h35m to fully train the model for 200 epochs on a single Titan RTX, while SCG-OL requires 108h40m, thus 8x slower. This is mostly due to the two-stage approach of SCG-OL which includes the non-differentiable localisation module and the more expensive activation function in the attention mechanism.

\subsection{Ablation studies}
\label{sec:exp_ablation}
We further analyse \newmname~to justify the usefulness of the commonsense relationships and our new attentional message passing mechanism. We also investigate the impact of increasing the number of message passing layers. To verify the applicability in 3D, we also evaluate the localisation performance of our method in comparison to the state-of-the-art methods. Lastly, we provide in-depth investigation on how the attention weights evolve over the message passing when forming the node and edge representation. 

\noindent\textbf{Which commonsense relationship is more important?} In order to better understand the effects of using different commonsense relationships, we compare the performance of \newmname~against four variants where the \newgraph~contains: i) only \emph{Proximity} edges without commonsense relationships, ii) \emph{Proximity} edges with \emph{AtLocation} edges, iii) \emph{Proximity} edges with \emph{UsedFor} edges, and vi) \emph{Proximity} edges with \emph{AtLocation} and \emph{UsedFor} edges.
We report the main Localisation Success Rate (LSR) measure for all variants, as well as the scene average percentage of object nodes which are linked by 0, 1, or 2 types of semantic edges, i.e. \emph{AtLocation} and \emph{UsedFor} edges.

\noindent\textbf{Discussion.} Table~\ref{tab:ablation:relation} shows that \emph{AtLocation} is more effective than \emph{UsedFor} for localising objects. This is reasonable, since the \textit{AtLocation} edge leads to message passing among objects that are connected in the same location, containing information more relevant to the localisation task. However, the best performance is obtained when the \newgraph~rely on all types of edges which provides a higher connectivity among object nodes to concept nodes. There are~80\% object nodes linked to concept nodes by both \textit{AtLocation} and \textit{UsedFor} edges, leading to a more effective knowledge fusion than when only one type of semantic edge is used.
\newline

\begin{table}[t]
\caption{Impacts of different ConceptNet relationships with the proposed \newmname. LSR: Localisation Success Rate.} 
\resizebox{1\linewidth}{!}{
\begin{tabular}{|c|P{0.18\linewidth}|P{0.18\linewidth}|P{0.18\linewidth}|c|}
\hline
\multirow{2}{*}{Edge Types} & \multicolumn{3}{|c|}{Obj. linked by \emph{n} semantic edges (\%)} & \multirow{2}{*}{LSR $\uparrow$} \\ \cline{2-4}
& 0 & 1 & 2 &\\ \hline \hline

Proximity  & 100 & 0 & 0 & 0.257 \\ \hline
{\it AtLocation}, Proximity & 8 & 92 & 0 & 0.292 \\\hline
{\it UsedFor}, Proximity & 19 & 81 & 0 & 0.272\\\hline
{\it AtLocation, UsedFor}, Proximity & 8 & 12 & 80 & \textbf{0.297} \\ \hline
\end{tabular}
} 
\label{tab:ablation:relation}
\end{table}

\noindent\textbf{Which attention network is more effective?}
We examine the usefulness of the proposed attention mechanism in \newmname~compared to other, commonly used, attentional message passing modules in the GNN literature. As most of these approaches do not support the use of edge features, we modify the node features for this ablation study to include the positional information to the node features. For a fair comparison, we remove the edge embedding from \newmname. The set of attention networks we compare with is listed below: 
\begin{itemize}[noitemsep,nolistsep]
\item  \textbf{No attention~\cite{NIPS2017_GraphSAGE}} is the first baseline, where we use GraphSAGE without relying on any attention module.

\item \textbf{GAT~\cite{velickovic2018graph}} adds an attention mechanism to the message passing procedure.

\item \textbf{GATv2~\cite{brody2021gatv2}} is similar to GAT but improves the attention mechanism in terms of the expressiveness and addresses the problem of "static attention" when using GATs for message passing.

\item \textbf{HAN~\cite{wang2019heterogeneous}} defines multiple meta-paths that connect neighbouring nodes either by specific node or edge types. It employs attentional message passing sequentially by first calculating the semantic-specific node embedding and then updating them by using an attention mechanism~\cite{vaswani2017attention}. With \newgraph~we define three sets of meta neighbours, i.e., the proximity neighbours, the \emph{AtLocation} neighbours, and the \emph{UsedFor} neighbours, connected by the specific edges. We implement the message passing for each meta-path using specialised GraphTransformer layers.

\item \textbf{GraphTransformer~\cite{shi-graphtransformer}} is similar to ours, except that it does not accommodate sparse attention and has less expressive power due to the smaller number of parameters. This module is essentially a porting of the scaled dot-product attention mechanism~\cite{vaswani2017attention} to GNNs. 
\end{itemize}

\noindent\textbf{Discussion.}
As shown in Table~\ref{tab:ablation:attention}, different attention modules can produce results that vary greatly in terms of LSR. Among all, HAN achieves the worst performance, showing that features are better to be propagated simultaneously rather than sequentially. GAT and GraphTransformer perform better than HAN, yet it is still worse than GraphSAGE which uses no attention. This is potentially due to the limitations of the standard attention mechanism when used in GNNs~\cite{brody2021gatv2}. GraphSAGE is a general inductive framework that leverages node feature information at different depths and is proven to work well on large graphs. In general the attention module should be carefully designed in order to provide advantageous performance. For example, GATv2 improves the localisation performance by fixing the static attention problem of the standard GAT. Our proposed attention mechanism achieves the best overall performance in terms of LSR, thanks to its increased expressive power and the ability to reason on sub-graphs during the message passing procedure, due to its capacity to learn sparse attention weights.\newline

\begin{table}[t]
\centering
\caption{Impacts of different attention modules for the object localisation task with our \newmname. LSR: Localisation Success Rate. %
}

\begin{tabularx}{1\linewidth}{|C|C|c|}
\hline
Attentional Network & LSR $\uparrow$ \\ 
\hline \hline
No attention \cite{NIPS2017_GraphSAGE} & 0.199 \\ \hline
GAT \cite{velickovic2018graph} & 0.179 \\ \hline
GATv2 \cite{brody2021gatv2}& 0.202 \\ \hline
HAN \cite{wang2019heterogeneous}& 0.137 \\ \hline 
GraphTransformer \cite{shi-graphtransformer} & 0.187 \\ \hline
Ours & \textbf{0.215} \\ \hline
\end{tabularx}\label{tab:ablation:attention}
\end{table}

\noindent\textbf{Does the number of message passing layers and the final node concatenation of \newmname~make a difference?}
\begin{table}[t]
\small
\centering
\caption{Impact of different numbers of message passing layers in our \newmname. LSR: Localisation Success Rate. }
\begin{tabularx}{0.9\linewidth}{|C|c|c|c|c|c|}
\hline
\# Layers & 1 & 2 & 3 & 4 & 5\\ \hline
LSR $\uparrow$ & 0.180 & 0.257 & 0.283 & \textbf{0.297} & 0.285 \\\hline
\end{tabularx}
\label{tab:ablation:msgpassing}
\end{table}

We examine a set of variants of our \newmname~with varying numbers of message passing layers ranged from 1 to 5. Table~\ref{tab:ablation:msgpassing} shows that using four message passing (MP) layers leads to the best performance. When using a single MP layer, there is not enough information regarding the context to be propagated to the nodes and this leads to the worst performance. With more than two MP layers, the performance starts to increase, saturating at four layers. With additional layers, we observe that the performance starts to degrade. This might be due to the over-smoothing problem~\cite{chen-oversmoothing-2020, Oono2020Graph}, where after multiple message passing rounds, the embeddings for different nodes are indistinguishable from one another. Given the best layer number, we also validate the choice of concatenating the original embedding to the aggregated \emph{contextual} ones, instead of using only the aggregated features. Concatenation is more advantageous with an LSR of $0.29$, while directly using the aggregated node representation leads to an LSR of $0.28$. We argue that this happens because concatenation allows the network to still remember the initial representation, developing a better understanding of the context after message passing.\newline

\begin{table}[t]
\small
\centering
\caption{Comparison of object localisation performance in the 3D environment instead of on the 2D floor plane.}
\begin{tabularx}{0.6\linewidth}{|C|c|}
\hline
 Method & LSR $\uparrow$ \\ \hline \hline
LayoutTransformer~\cite{gupta2021Layout}  & 0.158 \\ \hline
\mname~\cite{Giuliari2022Spatial}  & 0.048 \\ \hline
\newmname & \textbf{0.258} \\ \hline
\end{tabularx}
\label{tab:ablation:3d}
\end{table}

\begin{figure*}[th!]
    \centering
    \includegraphics[width=0.8\linewidth]{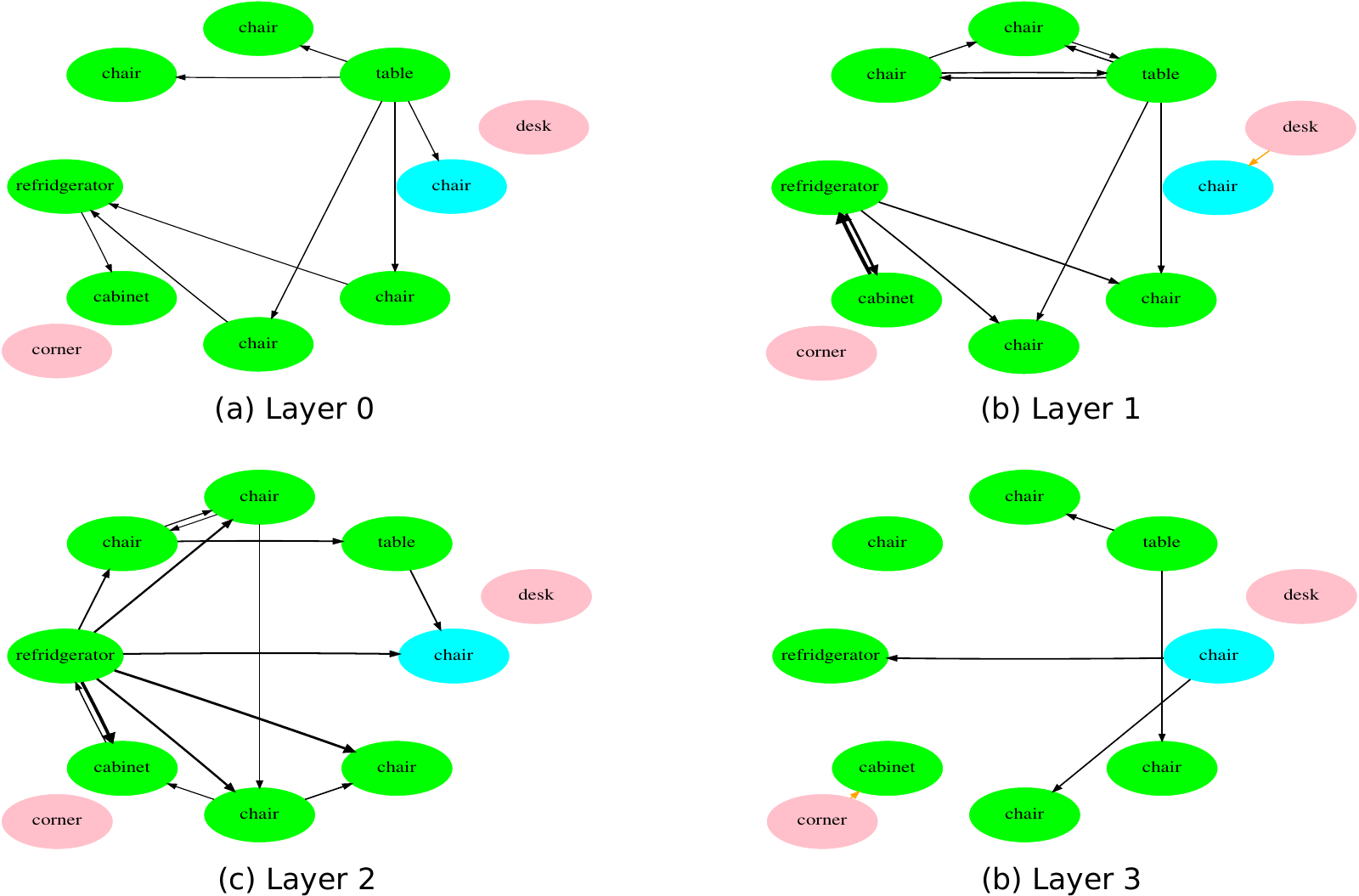}
    \caption{Feature propagation at different layers of our GNN that are directed by our attention module. The cyan node indicate the target object, the green nodes represent the scene nodes, and the pink nodes represent the concept nodes. The black edges indicates the sharing of information between two nodes in the direction indicated by the arrows. For ease of visualisation, we show edges with a mean attention weight over the heads that are superior to 0.2\%, and only display concept nodes that are connected via these type of edges.}
    \label{fig:attention_graph}

    \centering
    \includegraphics[width=0.7\linewidth]{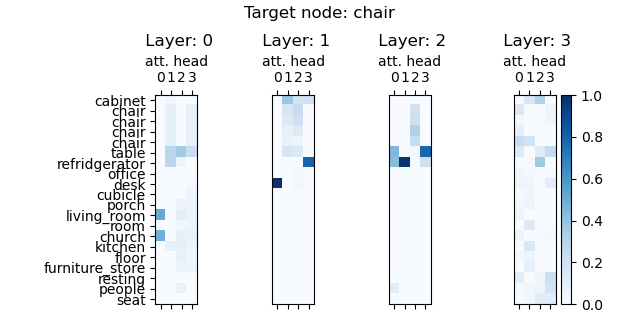}
    \caption{Attention weights for messages that are propagated to the target node are indicated in Fig.~\ref{fig:attention_graph}. The network learns to propagate information from different nodes by leveraging different attention heads. The first and last layer the network propagate information from most of the neighbouring nodes, while the intermediate layers focus on few specific nodes.}
    \label{fig:target_attention}
\end{figure*}

\noindent\textbf{{Localising in 3D}.} We examine the network capability to localise the target object directly in 3D scenes instead of on the 2D floor plane. We compare \newmname~with \mname~\cite{Giuliari2022Spatial} and LayoutTransformer~\cite{gupta2021Layout} with the following modifications:
\begin{itemize}
    \item \textbf{Layout Transformer~\cite{gupta2021Layout}}: We input to the model with the 3D coordinates $(x,y,z)$ of the objects instead of only $(x,y)$ and generate the corresponding 3D position of the target object.
    \item \textbf{\mname}~\cite{Giuliari2022Spatial}: We modify the input \oldgraph\ so that the pairwise distances are calculated using the 3D coordinates. The localisation module based on multilateration is changed to find the minima in the 3D space.
    \item \newmname: We modify the input \newgraph\ so that the relative positions are calculated in the 3D domain. The network is modified to predict the 3D relative positions.
\end{itemize}

Table~\ref{tab:ablation:3d} reports the localisation performance in the 3D scenes. We can observe that all the three methods suffer a drop in terms of LSR performance due to the increased difficulty level of the problem. \mname\ experiences the highest drop in performance, with an LSR score of only 0.05, down from its original score of 0.24 when evaluated in the 2D domain. This decrease in performance can be attributed to a difficulty in representing the object arrangement using only distances when an additional dimension is considered. %
Utilising a less abstract representation by using relative positions between objects leads to much more accurate results. Despite the increased problem difficulty, our proposed \newmname\ achieves the best LSR of~0.26, which is significantly higher than the second-best method LayoutTransformer with a LSR of 0.15. \newline

\noindent\textbf{{Attention Visualisation}} In Fig.~\ref{fig:attention_graph} and Fig.~\ref{fig:target_attention} we show how the network prioritises the exchange of information when localising a chair. Note that our network does not use the softmax function when calculating the attention weights, thus they do not necessarily sum to one. We normalise the weights for the visualisation results. Fig.~\ref{fig:attention_graph} shows the features propagated via message passing that are assigned a high weight by our attention modules. The network learns to operate very differently depending on the layer, and most of the attention weights are given to edges between object nodes. The network also learns to attend differently to instances of the same object based on the scene geometry that is described by the edge features. For instance, in the first layer only two of the five chairs nodes propagate their features with a high weight to the refrigerator. Incidentally, these nodes represent the two chairs closest to the fridge. Fig.~\ref{fig:target_attention} shows the different heads' attention scores for messages that are propagated to the target node. We can see that each head focuses on different nodes: some heads are giving high weights to specific nodes, e.g. head zero and three of the second layer, while others balance the features from many nodes, e.g. head two and three of the first layer. Lastly, we can see that most of the commonsense information is propagated in the first and last layer of the GNN.

\section{Conclusions}
\label{sec:conclusion}
We proposed an novel scene graph model, the \newgraph~, to address the problem of localising objects in a partial 3D scene. The spatial information regarding the arrangement of the object is described via directional edges with relative positions instead of undirectional relative distances as in the prior work. With the proposed \newgraph, we developed a new GNN-based solution for object localisation, \newmnamefull, that can directly estimate the position of the target object by predicting its relative positions with respect to other objects in the partially observed scene, leading to an efficient end-to-end trainable solution. Our approach also featured a novel attention module to further improve the localisation performance. We thoroughly evaluated our proposed method on the partial scene dataset and proved its superior performance in terms of localisation success rate against baselines and the state-of-the-art methods. Finally, we showed that our approach can be applied for 3D object localisation with a marginal performance drop, while the previous state-of-the-art method degrades dramatically due the increased localisation difficulty. Future work will focus on scaling our proposed approach to large-scale outdoor scenarios and extending to robotic applications.

\ifCLASSOPTIONcaptionsoff
  \newpage
\fi

\bibliographystyle{IEEEtran}
\bibliography{SCG}

\begin{IEEEbiography}[{\includegraphics[width=1in,height=1.2in,clip,keepaspectratio]{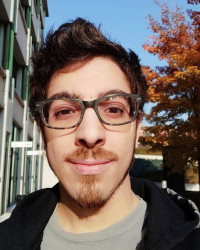}}]{Francesco Giuliari} (Student Member, IEEE)
is a PhD student at University of Genoa. He is currently affiliated with Istituto Italiano di Tecnologia under the supervision of Dr. Alessio Del Bue. He received his MsC in Computer Science from University of Verona in 2018. His main research interests are in computer vision, scene understanding and vision-based agent navigation.
\end{IEEEbiography}
\vspace*{-2\baselineskip}
\begin{IEEEbiography}[{\includegraphics[width=1in,height=1.2in,clip,keepaspectratio]{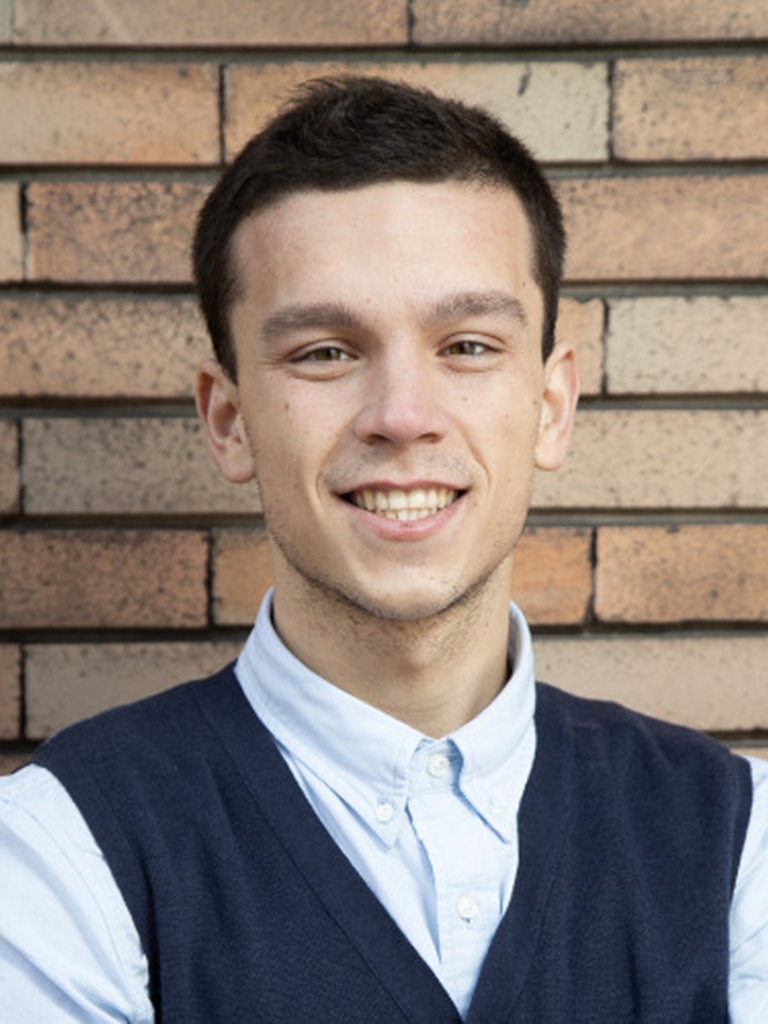}}]{Geri Skenderi}
is a PhD student at the University of Verona, working under the supervision of Prof. Marco Cristani. His research interests cover the broad area of deep learning, with an applicative focus on forecasting. Before his PhD, he received his Master’s degree in Computational Data Science at the Free University of Bolzano-Bozen in 2020.
\end{IEEEbiography}

\vspace*{-2\baselineskip}
\begin{IEEEbiography}[{\includegraphics[width=1in,height=1.1in,clip,keepaspectratio]{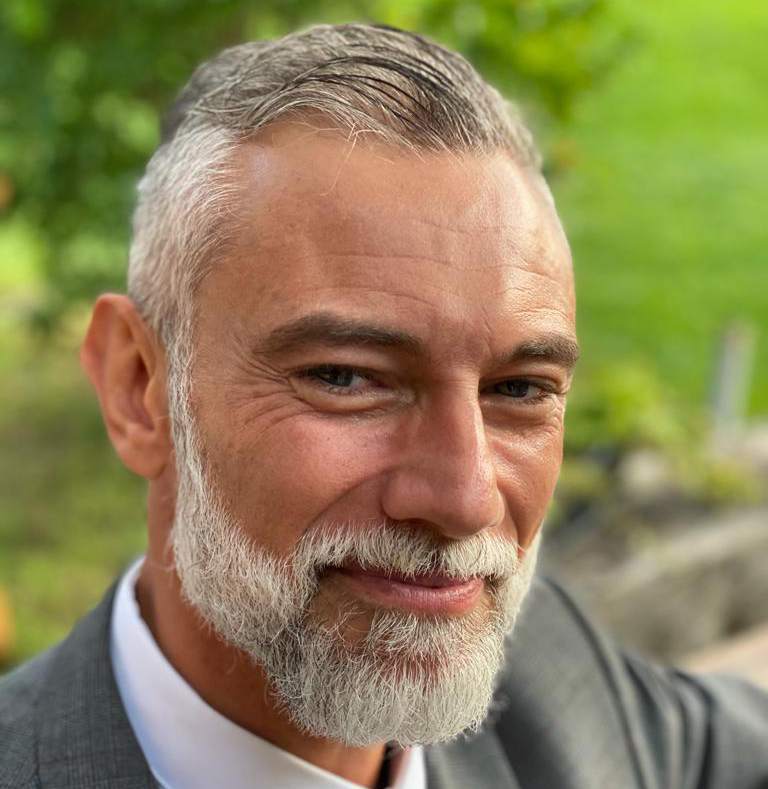}}]{Marco Cristani} is Full Professor (Professore Ordinario) at the Computer Science Department, University of Verona,
Associate Member at the National Research Council (CNR), External Collaborator at the Italian Institute of Technology (IIT). His main research interests are in statistical pattern recognition and computer vision, mainly in deep learning and generative modeling, with application to social signal processing and fashion modeling. On these topics he has published more than 180 papers, including two edited volumes, 46 international journal papers, 126 conference papers and 13 book chapters. He has organized 11 international workshops, cofounded a spin-off company, Humatics, dealing with e-commerce for fashion. He is or has been Principal Investigator of several national and international projects, including PRIN and H2020 projects. He is a member of the editorial board of the Pattern Recognition and Pattern Recognition
Letters journals. He is Managing Director of the Computer Science Park, a technology transfer center at the University of Verona. He is a member of ACM, IEEE and IAPR.
\end{IEEEbiography}

\vspace*{-2\baselineskip}
\begin{IEEEbiography}[{\includegraphics[width=1in,height=1.25in,clip,keepaspectratio]{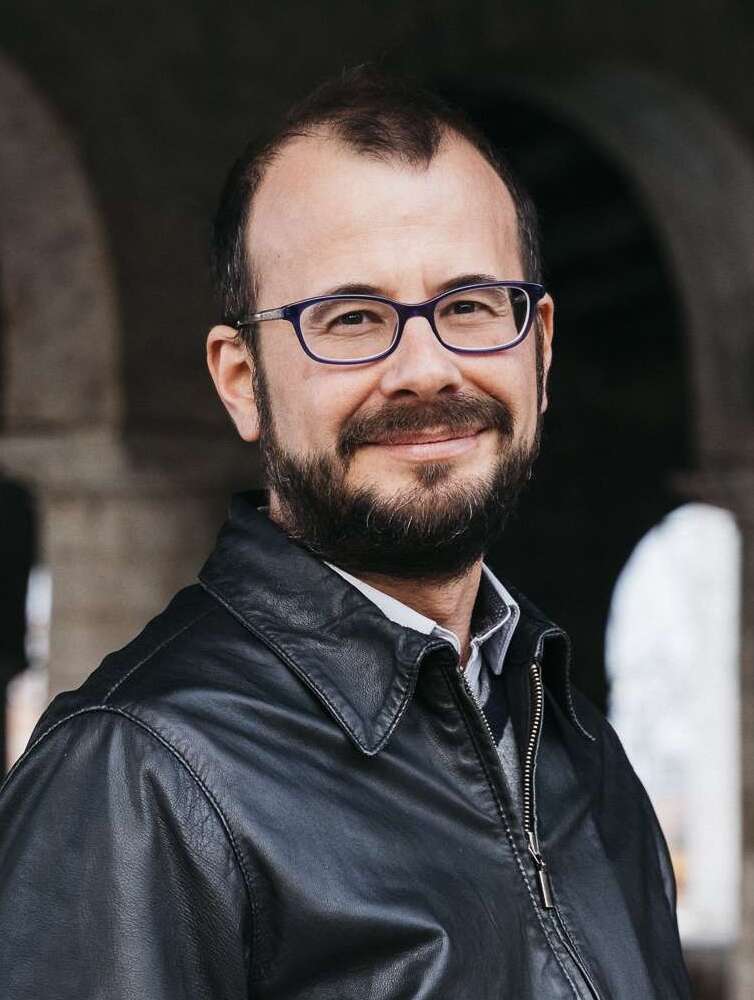}}]{Alessio Del Bue} (Member, IEEE) is a Tenured Senior Researcher leading the Pattern Analyisis and computer VISion (PAVIS) Research Line of the Italian Institute of Technology (IIT), Genoa, Italy. He is a coauthor of more than 100 scientific publications in refereed journals and international conferences on computer vision and machine learning topics. His current research interests include 3D scene understanding from multi-modal input (images, depth, and audio) to support the development of assistive artificial intelligence systems. He is a
member of the technical committees of major computer vision conferences (CVPR, ICCV, ECCV, and BMVC). He serves as an Associate Editor for Pattern Recognition and Computer Vision and Image Understanding
journals. He is a member of ELLIS.
\end{IEEEbiography}
\vspace*{-2\baselineskip}
\begin{IEEEbiography}[{\includegraphics[width=1in,height=1.25in,clip,keepaspectratio]{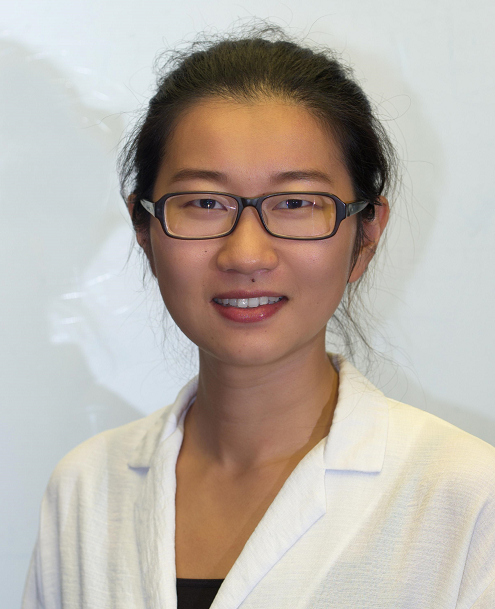}}]{Yiming Wang}
is a researcher in the Deep Visual Learning (DVL) unit at Fondazione Bruno Kessler (FBK). Her research mainly focuses on vision-based scene understanding that facilitates automation for social good. Yiming obtained her PhD in Electric Engineering from Queen Mary University of London (UK) in 2018, working on vision-based multi-agent navigation. Since 2018, she has worked as a post-doc researcher in the Pattern Analysis and Computer Vision (PAVIS) research line at Istituto Italiano di Tecnologia (IIT), working on topics related to active 3D vision. She is actively serving as reviewers for top-tier conferences and journals in both the Computer Vision and Robotics domains.
\end{IEEEbiography}

\end{document}